\documentclass[conference]{IEEEtran}
\IEEEoverridecommandlockouts
\usepackage{booktabs}
\usepackage[noadjust]{cite}
\usepackage{amsmath,amssymb,amsfonts}
\usepackage{algorithmic}
\usepackage{graphicx}
\usepackage{overpic}
\usepackage{textcomp}
\usepackage{xcolor}
\usepackage{caption}
\usepackage{float}
\usepackage{array}
\usepackage{graphicx}
\usepackage{tikz}
\usepackage{capt-of}
\usetikzlibrary{calc}

\newcommand{\resLen}{0.135\textwidth}

\def\BibTeX{{\rm B\kern-.05em{\sc i\kern-.025em b}\kern-.08em
    T\kern-.1667em\lower.7ex\hbox{E}\kern-.125emX}}
\begin{document}

\title{Scene Perceived Image Perceptual Score (SPIPS): combining global and local perception for image quality assessment\\
}

\author{\IEEEauthorblockN{Zhiqiang Lao, Heather Yu}
\IEEEauthorblockA{
\textit{Futurewei Technologies Inc}\\
Basking Ridge, New Jersey, USA \\
\{zlao,hyu\}@futurewei.com}
}

\IEEEpubid{\makebox[\columnwidth]{978-1-xxxx-xxxx-x/25/\$31.00~\copyright~2025 IEEE \hfill} \hspace{\columnsep}\makebox[\columnwidth]{ }}

\maketitle
\begin{figure*}[t]
    \centering
    \begin{minipage}[t]{0.49\linewidth}
        \centering

        \begin{tikzpicture}
            \node[inner sep=0pt] (img) at (0,0) {\includegraphics[width=\linewidth]{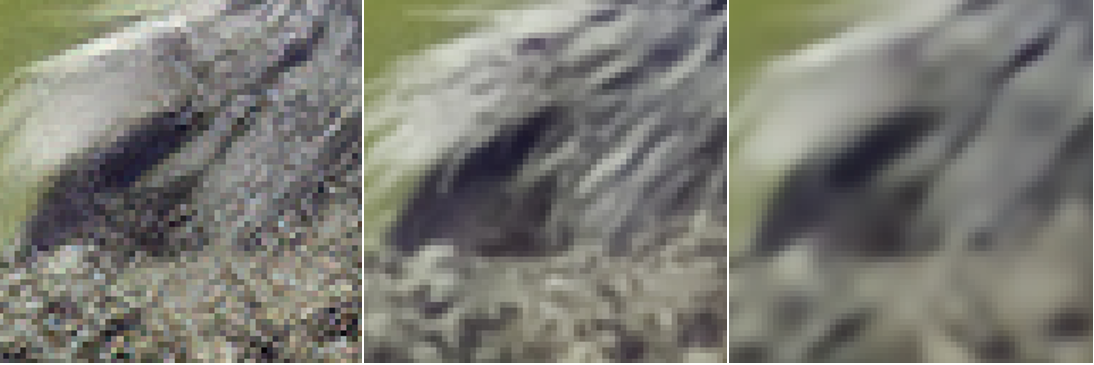}};
            \path let \p1 = ($(img.south east) - (img.south west)$),
                      \p2 = ($(img.north west) - (img.south west)$) in
            node[text=white, font=\bfseries\large] at ($(img.south west) + (0.166*\x1, 0.9*\y2)$) {image0}
            node[text=white, font=\bfseries\large] at ($(img.south west) + (0.5*\x1, 0.9*\y2)$) {reference}
            node[text=white, font=\bfseries\large] at ($(img.south west) + (0.833*\x1, 0.9*\y2)$) {image1};
        \end{tikzpicture}

        \small
        \vspace{6pt}
        \begin{tabular}{m{0.7cm}*{2}{m{0.65cm}}m{0.5cm}*{4}{m{0.65cm}}}
            & PSNR & SSIM & VIF & LPIPS & DISTS & SPIPS & Human \\
            \hline
            \vspace{6pt}
            0$>$1 &  &  &  &  &  & \textcolor{green}{$\checkmark$} & \textcolor{green}{$\checkmark$} \\
            1$>$0 & \textcolor{red}{$\checkmark$} & \textcolor{red}{$\checkmark$} & \textcolor{red}{$\checkmark$} & \textcolor{red}{$\checkmark$} & \textcolor{red}{$\checkmark$} &  &  \\
        \end{tabular} \\
        \vspace{6pt}
        \small (a) \textbf{Human Preference}: $image0 > reference > image1$
    \end{minipage}
    \hfill
    \begin{minipage}[t]{0.49\linewidth}
        \centering

        \begin{tikzpicture}
            \node[inner sep=0pt] (img) at (0,0) {\includegraphics[width=\linewidth]{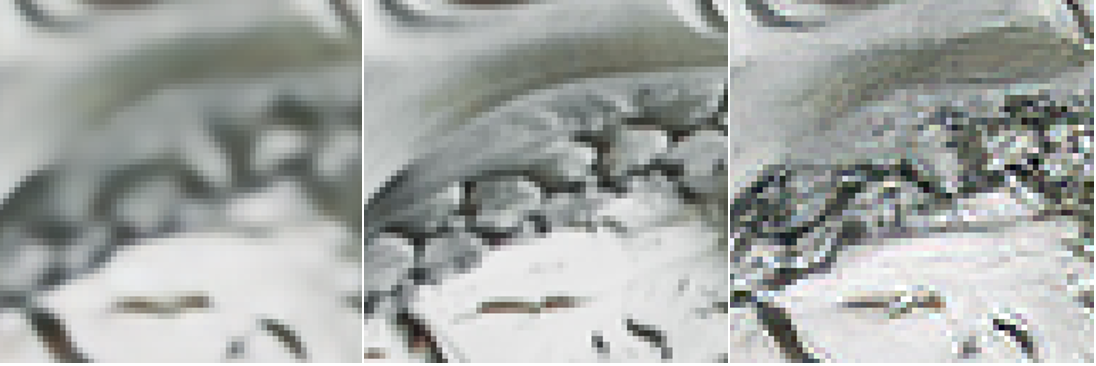}};
            \path let \p1 = ($(img.south east) - (img.south west)$),
                      \p2 = ($(img.north west) - (img.south west)$) in
            node[text=white, font=\bfseries\large] at ($(img.south west) + (0.166*\x1, 0.9*\y2)$) {image0}
            node[text=white, font=\bfseries\large] at ($(img.south west) + (0.5*\x1, 0.9*\y2)$) {reference}
            node[text=white, font=\bfseries\large] at ($(img.south west) + (0.833*\x1, 0.9*\y2)$) {image1};
        \end{tikzpicture}

        \small
        \vspace{6pt}
        \begin{tabular}{m{0.7cm}*{2}{m{0.65cm}}m{0.5cm}*{4}{m{0.65cm}}}
            & PSNR & SSIM & VIF & LPIPS & DISTS & SPIPS & Human \\
            \hline
            \vspace{6pt}
            0$>$1 & \textcolor{red}{$\checkmark$} & \textcolor{red}{$\checkmark$} & \textcolor{red}{$\checkmark$} & \textcolor{red}{$\checkmark$} & \textcolor{red}{$\checkmark$} &  & \\
            1$>$0 &  &  &  &  &  & \textcolor{green}{$\checkmark$} & \textcolor{green}{$\checkmark$} \\
        \end{tabular} \\
        \vspace{6pt}
        \small (b) \textbf{Human Preference}: $image0 < reference < image1$
    \end{minipage}

    \caption{Qualitative comparison of different metrics against human preference on the BAPPS dataset. SPIPS consistently aligns with human judgment across different distortion types.}
    \label{fig:visual_compare}
\end{figure*}

\begin{abstract}
The rapid advancement of artificial intelligence and widespread use of smartphones have resulted in an exponential growth of image data, both real (camera-captured) and virtual (AI-generated). This surge underscores the critical need for robust image quality assessment (IQA) methods that accurately reflect human visual perception. Traditional IQA techniques primarily rely on spatial features—such as signal-to-noise ratio, local structural distortions, and texture inconsistencies—to identify artifacts. While effective for unprocessed or conventionally altered images, these methods fall short in the context of modern image post-processing powered by deep neural networks (DNNs). The rise of DNN-based models for image generation, enhancement, and restoration has significantly improved visual quality, yet made accurate assessment increasingly complex.

To address this, we propose a novel IQA approach that bridges the gap between deep learning methods and human perception. Our model disentangles deep features into high-level semantic information and low-level perceptual details, treating each stream separately. These features are then combined with conventional IQA metrics to provide a more comprehensive evaluation framework. This hybrid design enables the model to assess both global context and intricate image details, better reflecting the human visual process, which first interprets overall structure before attending to fine-grained elements. The final stage employs a multilayer perceptron (MLP) to map the integrated features into a concise quality score. Experimental results demonstrate that our method achieves improved consistency with human perceptual judgments compared to existing IQA models.
\end{abstract}


\section{Introduction}
The effectiveness of computer vision in real-world applications depends on how closely the target function aligns with the human visual system. End-to-end solutions for tasks like denoising, super-resolution, and lossy compression, etc. \cite{balle2016end,balle2018variational} rely on differentiable similarity metrics that accurately reflect human perception of visual changes. However, commonly used metrics that assess pixel-level differences, such as PSNR and MSE, while differentiable, fail to align well with human visual perception.

The inability of pixel-level metrics to accurately capture human perception has led to the development of patch-level similarity metrics, influenced by the psychophysics subfield of human psychology. The most effective so far is the multi-scale structural similarity metric (MS-SSIM)\cite{wang2003multiscale,wang2004image}, which accounts for luminance and contrast perception. However, despite these advancements, the complexity of the human visual system remains challenging to model manually. This is evident in MS-SSIM’s shortcomings in predicting human preferences in standardized image quality assessment (IQA) experiments\cite{zhang2018unreasonable}.

To move beyond manually designed similarity metrics, researchers have adopted deep features extracted from large pre-trained neural networks. The Learned Perceptual Image Patch Similarity (LPIPS)\cite{zhang2018unreasonable} metric, for instance, relies on the L2 distance between these deep features to approximate human perception. In the same study, the authors introduced the Berkeley Adobe Perceptual Patch Similarity (BAPPS) dataset, which has since become a widely recognized benchmark for evaluating the perceptual consistency of similarity metrics. LPIPS leverages deep features as input for a smaller neural network trained on human-annotated data from BAPPS, which captures human preferences for image similarity\cite{kumar2022better}.

The human visual system exhibits a strong hierarchical perception ability, typically employing a top-down approach—first grasping the overall scene before focusing on details. In contrast, due to the limitations of visual sensors like cameras, images are stored in computers as pixel arrays, leading computer vision to adopt a bottom-up approach that prioritizes details before forming a holistic understanding. As a result, image quality assessment in computer vision often emphasizes fine details rather than the broader image context. Learning-based image quality assessment methods often utilize backbone networks such as AlexNet or VGG, which offer high-level perceptual capabilities. However, these models were originally developed for tasks like image classification, object detection, and segmentation - all of which prioritize recognizing salient content over assessing visual quality. Consequently, the features they extract are not inherently suited to the goal of image quality assessment. Relying on features optimized for classification to evaluate image quality introduces a mismatch that can compromise performance. This underscores a key limitation of current learning-based methods: without task-specific adaptation, classification-driven features may overlook critical aspects of perceptual quality. Recognizing these limitations, this paper enhances the role of global features in image quality assessment by integrating both deep and traditional features. The proposed metric, guided by this principle, achieves closer alignment with human visual perception.

Our contributions can be summarized as:  
\begin{itemize}
\item High-level semantic and low-level perceptual image features are treated distinctly, processed independently, and then fused. This differentiated processing enables the model to consider both fine-grained details and overall scene understanding when assessing image quality, resulting in a more comprehensive evaluation that aligns more closely with human visual perception.  
\item Integrating traditional image quality assessment techniques with deep learning models helps reduce the common inconsistencies in their evaluation outcomes.
\end{itemize}

\section{Related works}
In this section we review closely related literature for data-free and learned (both unsupervised and supervised) full-reference image quality assessment (FR-IQA).

\subsection{Data-free FR-IQA}
Data-free distortion metrics operating at the pixel level such as mean squared error (MSE) are commonly used in lossy compression applications \cite{cover1999elements} but have long been known to correlate poorly with human perception \cite{girod1993sGirod}. Patch-level metrics have been shown to correlate better with human judgement on psychophysical tasks. Most notably, the Structural Similarity Index (SSIM \cite{wang2004image}, as well as its multi-scale variate MS-SSIM \cite{wang2003multiscale}, compare high level patch features such as luminance and contrast to define a distance between images \cite{wang2004image}. SSIM \cite{wang2004image} is widely used in commercial television applications, and MS-SSIM \cite{wang2003multiscale} is a standard metric for assessing performance on many computer vision tasks. The method presented in this work is also data-free and outperforms MS-SSIM on benchmark datasets.

The PSNR calculates the squared discrepancies between a reference and a generated image, offering a straightforward measure of image degradation \cite{talebi2018learned}. Visual Information Fidelity (VIF) is another metric that quantifies the amount of visual information preserved in an image compared to a reference image, based on natural scene statistics and the characteristics of the human visual system \cite{sheikh2006image}. SSIMPLUS, introduced by Rehman et al. \cite{rehman2015display}, advances SSIM by integrating factors like human vision, display characteristics, and viewing conditions by enabling real-time perceptual quality predictions. There are some successful FR methods based on information theoretic models such as Information Fidelity Criterion (IFC) \cite{Sheikh2005}.

\subsection{Learned FR-IQA}

On the model-based front, the FID evaluates the statistical distance between feature vectors of real and generated images captured by the Inception model serving as an indicator of visual fidelity \cite{heusel2017gans}. The Inception Score (IS), developed by Salimans et al. \cite{salimans2016improved} quantifies image diversity and clarity using the predictive entropy of a classifier trained on diverse datasets like ImageNet. Kernel Inception Distance (KID), introduced by Binkowski et al. \cite{binkowski2018demystifying}, improves upon FID by employing a non-parametric approach using a polynomial kernel to compute the squared Maximum Mean Discrepancy between feature distributions. This method avoids assumptions about the distributional form of activations, better accommodating the non-negative nature of ReLU activations in deep networks.

Many learned FR-IQA methods are designed mirroring the learned perceptual image patch similarity (LPIPS) method of \cite{zhang2018unreasonable} and Deep Image Structure and Texture Similarity (DISTS) \cite{ding2020image}, where a neural network is trained on some auxiliary task and the intermediate layers are taken as perceptual representations of an input image. An unsupervised distance between images is defined as the L2 norm of the difference between their representations. A supervised distance uses the representations as inputs to a second model that is trained on human annotated data regarding the perceptual quality of the input images (e.g., labels of 2-AFC datasets discussed in Section 2). Taking representations from neural networks that perform well on their auxiliary task does not guarantee good performance on perceptual tasks \cite{kumar2022better}, making it difficult to decide which existing models will yield perceptually relevant distance functions. In \cite{zhou2019visual}, the Structure-Texture Decomposition (STD) is used to measure the structural, textural and high-frequency similarities. Zhang et al. \cite{zhang2023perception} proposed a perception-driven similarity-clarity tradeoff to balance quality scores between the referenced similarity and the no-reference clarity, and well cope with the visually pleasing false textures generated by GAN-based SR algorithms.

Self-supervision was used by Madhusudana et al. \cite{madhusudana2023conviqt,madhusudana2022image} and Wei et al. \cite{wei2022contrastive} for unsupervised and supervised FR-IQA. Images are corrupted with pre-defined distortion functions and a neural network is trained with a contrastive pairwise loss to predict the distortion type and degree. The unsupervised distance is defined as discussed previously and ridge regression is used to learn a supervised distance function. This method requires training data, while our method requires no training at all.

Benefiting from the powerful feature expression ability, deep-learning-based generic approaches including CNN-based models \cite{kang2014convolutional,madhusudana2022image,pan2022dacnn,zhou2023blind} and Transformer-based models \cite{golestaneh2022no,yang2022maniqa,qin2023data,sun2023blind} have also shown excellent performance for image distortions. They put great efforts into the challenges of the limited size of train samples and complex distortion conditions. In \cite{pan2022dacnn}, the feature extraction network was trained in a distortion aware manner to cope with various distortions. Zhou et al. \cite{zhou2023blind} designed a self-supervised architecture to enhance the representation ability to content and distortion.

\section{Proposed method}
\begin{figure*}[t]
    \centering
    \includegraphics[width=0.7\linewidth]{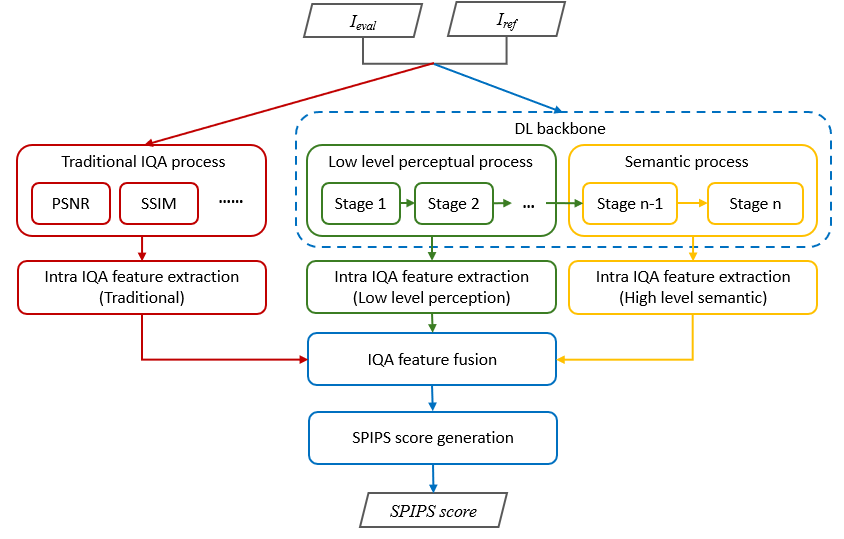}
    \caption{\textbf{The overall framework} of our SPIPS model is structured into three modules: the traditional image quality assessment module (red), the low-level image perception feature assessment module (green), and the high-level image semantic feature assessment module (yellow). SPIPS takes two input images, \(I_{eval}\) (the image to be evaluated) and \(I_{ref}\) (the reference or ground truth image). The traditional image quality assessment module (red) analyzes quality differences between corresponding regions of \(I_{eval}\) and \(I_{ref}\) using standard industry metrics such as PSNR, SSIM, and MS-SSIM, producing a quality assessment map. Both \(I_{eval}\) and \(I_{ref}\) are then fed into a pre-trained deep learning backbone (e.g., AlexNet, VGG, SqueezeNet, represented by the light blue dotted line) to extract feature maps from each layer before the fully connected layers. These feature maps are categorized based on their focus: low-level image perception features (derived from all layers except the last two) and high-level semantic features (from the final two layers of the feature stack). The mean squared error (MSE) between the deep feature representations of \(I_{eval}\) and \(I_{ref}\) is then computed to generate a deep evaluation map. Each evaluation map undergoes independent quality enhancement before being combined into the final image quality score of \(I_{eval}\) through a weighted averaging process. Similar to the optimization approach in the LPIPS model \cite{zhang2018unreasonable}, this score is compared to human visual perception scores to compute a loss value. The SPIPS model parameters are iteratively optimized via backpropagation, following a similar training strategy to that used in the LPIPS model.
    }
    \label{fig:framework}
\end{figure*}

Figure \ref{fig:framework} illustrates the overall framework of our SPIPS image quality assessment model, which is composed of three key modules:  

\vspace{2pt}
\noindent\textbf{Traditional Image Quality Assessment (IQA) Module}: Given a pair of input images, this module computes traditional image quality metrics such as PSNR, SSIM, and MS-SSIM. Unlike conventional approaches that output a single numerical value, this module generates an assessment image of the same size as the input, providing a spatially aware quality evaluation.  

\vspace{2pt}
\noindent\textbf{Deep Feature-Based IQA Module}: Utilizing a pre-trained deep learning model (e.g., AlexNet, VGG), this module extracts features from multiple layers. Since different layers capture distinct aspects of an image, these features are categorized into two groups: low-level image perception features and high-level semantic features. For instance, in AlexNet, features from the first three layers are classified as perceptual features, while the last two layers are designated as semantic features. For a given pair of input images, their respective perceptual and semantic features are computed, and the mean squared error (MSE) between corresponding layers serves as the quality assessment result, quantifying the difference between the evaluated image and the reference image.  

\vspace{2pt}
\noindent\textbf{Image Quality Feature Extraction Module}: This module extracts quality-related features from the outputs of the previous two modules. These extracted features, which are directly relevant to image quality, are then fused to generate the final image quality score.

\subsection{Traditional IQA Module}\label{AA}
The traditional IQA module evaluates the quality of the image to be assessed using industry-standard formulas, including PSNR, SSIM, and MS-SSIM \cite{gonzalez2009digital,wang2004image, hore2010image,wang2003multiscale}. To align with the deep image quality assessment module while differing from conventional usage, two key modifications were made:  

\begin{itemize}
    \setlength\itemsep{0pt}
    \item Instead of producing a single numerical score, this module outputs a quality assessment value for each pixel, resulting in a quality assessment map rather than an averaged value.
    \item Typically, PSNR, SSIM, and MS-SSIM  are defined such that higher values indicate better image quality. However, in the deep feature-based module, image quality is assessed by measuring the difference in feature representations between the evaluated image and the reference image, where smaller differences indicate higher quality. To maintain consistency, the PSNR, SSIM, and MS-SSIM assessment maps are first normalized to the range \([0,1]\), and then each pixel’s value is subtracted from 1. This transformation ensures that lower values on the final assessment map correspond to higher image quality.
\end{itemize}

Given a pair of images, $I_{eval}\in\mathbb{R}^{3\times\mathrm{H}\times\mathrm{W}}$ (the image to be evaluated) and $I_{ref}\in\mathbb{R}^{3\times\mathrm{H}\times\mathrm{W}}$ (the ground truth), PSNR, SSIM, and MS-SSIM calculations are applied to assess their quality differences.
\begin{align}
    Q_{psnr} = 1 - \mathcal{N}(\mathcal{P}(I_{eval}, I_{ref})) \\
    Q_{ssim} = 1 - \mathcal{N}(\mathcal{S}(I_{eval}, I_{ref})) \\
    Q_{msssim} = 1 - \mathcal{N}(\mathcal{S}_{\text{MS}}(I_{eval}, I_{ref}))
\end{align}

where $Q_{psnr}\in\mathbb{R}^{3\times\mathrm{H}\times\mathrm{W}}$, $Q_{ssim}\in\mathbb{R}^{3\times\mathrm{H}\times\mathrm{W}}$, $Q_{msssim}\in\mathbb{R}^{\mathrm{C}\times\mathrm{H}\times\mathrm{W}}$, $\mathrm{C}$ is the number of scales, and $\mathcal{P}(\cdot)$ is the PSNR operation, $\mathcal{S}(\cdot)$ is the SSIM operation, $\mathcal{S}_{\text{MS}}(\cdot)$ is the MS-SSIM operation and $\mathcal{N}(\cdot)$ is the normalization to $[0,1]$ operation.

\subsection{Deep Feature-Based IQA Module}
Convolutional neural network (CNN)-based deep learning models possess strong feature extraction capabilities and have been widely applied in computer vision, including image quality assessment. Our SPIPS model leverages these powerful image features as well. CNN-based models typically extract features at different resolutions, where high-resolution features emphasize image details, while low-resolution features capture broader semantic characteristics.  

Like similar approaches, SPIPS utilizes pre-trained CNN models such as AlexNet, VGG, and SqueezeNet for feature extraction. However, unlike previous methods, SPIPS places greater emphasis on semantic features. In addition to extracting semantic information, it also captures features specifically related to image quality. These features are then integrated with perceptual features from image details and traditional image quality metrics, resulting in a more comprehensive visual assessment that aligns more closely with human visual perception.

Lets consider a pair of images, $I_{eval}\in\mathbb{R}^{3\times\mathrm{H}\times\mathrm{W}}$ (the image to be evaluated) and $I_{ref}\in\mathbb{R}^{3\times\mathrm{H}\times\mathrm{W}}$ (the ground truth), CNN-based feature extraction is as follows.
\begin{align}
    F_{eval} = \left\{ \Phi_{\text{CNN}}^{(l)}(I_{eval}) \mid l = 1, \dots, L \right\} \\
    F_{ref} = \left\{ \Phi_{\text{CNN}}^{(l)}(I_{ref}) \mid l = 1, \dots, L \right\}
\end{align}
where $F^{(l)}_{eval}$ , $F^{(l)}_{ref}\in\mathbb{R}^{\text{C}^{(l)}\times\mathrm{H^{(l)}}\times\mathrm{W^{(l)}}}$,  $C^{(l)}, H^{(l)}, W^{(l)}$ are the number of feature channels, height and weight at layer $l$ respectively, $\text{CNN}$ is pretrained backbone network (AlexNet, VGG, etc.), $L$ is the total number of feature extraction layers in $\text{CNN}$. 

The $\text{CNN}$ features based image quality maps can be computed as follows.
\begin{equation}
    Q_{\text{CNN}} = \left\{ (F^{(l)}_{eval} - F^{(l)}_{ref}) \odot (F^{(l)}_{eval} - F^{(l)}_{ref}) \mid l = 1, \dots, L \right\}
\end{equation}
where $Q^{(l)}_{\text{CNN}}\in\mathbb{R}^{\text{C}^{(l)}\times{\text{H}^{(l)}}\times{\text{W}^{(l)}}}$, \( \odot \) denotes the \textit{Hadamard product} \cite{styan1973hadamard} (element-wise multiplication), the operation computes the squared difference for each corresponding element of $F^{(l)}_{eval}$ and $F^{(l)}_{ref}$.

Based on the quality assessment focus of \( Q_{\text{CNN}} \) generated at different layers, \( Q_{\text{CNN}} \) is categorized into the image quality perception group and the image quality semantic group, defined as follows:
\begin{align}
    Q_{percept} = \left\{ Q^{(l)}_{\text{CNN}} \mid l = 1, \dots, L-2 \right\} \\
    Q_{semantic} = \left\{ Q^{(l)}_{\text{CNN}} \mid l = L-1, L \right\}
\end{align}
Using AlexNet as an example, which consists of five feature extraction layers ($L=5$), substituting $\text{CNN}$ with $\text{AlexNet}$ in equations above yields: $Q_{percept} = \left\{ Q^{(l)}_{\text{AlexNet}} \mid l = 1, 2, 3 \right\}$ and $Q_{semantic} = \left\{ Q^{(l)}_{\text{AlexNet}} \mid l = 4, 5 \right\}$

\subsection{Image Quality Feature Extraction Module}
At this stage, we have obtained three sets of image quality assessment maps: \( Q_{tradition} = \{ Q_{psnr}, Q_{ssim}, Q_{msssim} \} \), \( Q_{percept} \), and \( Q_{semantic} \). Next, we need to extract features directly related to image quality from these quality assessment maps. Similar to image feature extraction, the process of extracting image quality features also employs convolution operations. Each of the 3 image quality sets undergoes distinct convolution processes to ensure that the extracted image quality features effectively capture the unique characteristics of each group.

\begin{align}
    F_{tradition} = \mathcal{R}(\mathcal{C}_{tradition}(Q_{tradition}))   \\
    F_{percept} = \mathcal{R}(\mathcal{C}_{percept}(Q_{percept}))   \\
    F_{semantic} = \mathcal{R}(\mathcal{C}_{semantic}(Q_{semantic}))
\end{align}
where $\mathcal{C}_{tradition}(\cdot)$, $\mathcal{C}_{percept}(\cdot)$ and $\mathcal{C}_{semantic}(\cdot)$ are convolution operations apply to $Q_{tradition}$, $Q_{percept}$ and $Q_{semantic}$ respectively, $\mathcal{R}(\cdot)$ is ReLU.

\subsection{IQA feature fusion and score computation}

The final image quality assessment score is derived by combining the extracted image quality features to achieve a more comprehensive evaluation, represented as follows:

\begin{equation}
    score_{spips} = \lambda_{1}\bar{F}_{tradition} + \lambda_{2}\bar{F}_{percept} + \lambda_{3}\bar{F}_{semantic}
\end{equation}
where $\lambda_{1}$, $\lambda_{2}$ and $\lambda_{3}$ are weighting terms for tradition, percept and semantic respectively, determining their contribution in final $score_{spips}$ computation, and $\lambda_{1} + \lambda_{2} + \lambda_{3} = 1$. $\bar{F}_{tradition}$ is the mean value of $F_{tradition}$, so are $\bar{F}_{percept}$ and $\bar{F}_{semantic}$.

The $score_{spips}$ represents the final image quality evaluation for \( I_{eval} \), where a lower value indicates better quality.

\section{experiment}

\subsection{Datasets and experiments setup}\label{DAES}
BAPPS (Berkeley Adobe Perceptual Patch Similarity) dataset \cite{zhang2018unreasonable} is used to train and validate SPIPS model. This dataset is specifically designed to analyze the differences between image quality assessments performed by computer vision models and those perceived by the human visual system. It includes a diverse set of artifacts resulting from natural image processing, traditional algorithms (e.g., image compression), and CNN-based models. These artifacts are categoriezed into six groups: traditional distortions (\textit{Trad}), CNN-based distortions (\textit{CNN}), video deblurring (\textit{Deblur}), frame interpolation (\textit{Interp}), super-resolution (\textit{SR}), and colorization (\textit{Color}). Table \ref{tab:BAPPS} summarizes key statistics of the BAPPS dataset. The LPIPS model, which is built upon this dataset, is one of the most widely used deep feature-based image quality assessment models. The images included in the dataset have a resolution of 256 × 256. 

The BAPPS dataset uses two methods to evaluate image quality. The first, known as \textit{2AFC} (two-alternative forced choice), presents each sample with three images: $image0$, $reference$, and $image1$. The task is to decide which of the two images ($image0$ or $image1$) is closer in quality to the reference.  
The second method, called \textit{JND} (just noticeable differences), presents a pair of images for each sample and asks whether a noticeable difference exists between them.

\begin{table}[H]
    \centering
    \begin{tabular}{c|c}
        Properties & Value \\
        \hline
        \# of Ref. Images & 187.7k \\
        \# of Test Images & 375.4k \\
        Distortion / Enhancement Type & Simulated / DNN-based \\
        Image resolution & 256 × 256 \\
        \# of Human Annotations & 484.3k
    \end{tabular}
    \captionsetup{format=plain, labelsep=space}
    \captionsetup{font=footnotesize}
    \caption{Key statistics of BAPPS dataset.}
    \label{tab:BAPPS}
\end{table}
The commonly used correlation metrics, including Spearman’s Rank Correlation Coefficient (SRCC) \cite{sedgwick2014spearman}, Pearson’s Linear Correlation Coefficient (PLCC) \cite{sedgwick2012pearson}, and Kendall’s Rank Correlation Coefficient (KRCC) \cite{abdi2007kendall}, are utilized to evaluate the consistency between computer-based image quality assessments and human perception. Each of these metrics provides a different perspective on the evaluation of prediction performance.  

SRCC is a rank-based non-parametric metric that measures the monotonic relationship between predicted and human-labeled image quality scores, making it particularly effective for assessing relative ranking consistency rather than absolute differences. PLCC, on the other hand, evaluates the linear correlation between predicted and ground-truth scores, reflecting the accuracy of numerical predictions and how well they align with human judgment. Finally, KRCC quantifies the similarity between the ranking of predicted values and the ranking of human-provided labels, providing another perspective on prediction reliability.  

By utilizing these three correlation indicators together, a comprehensive evaluation of the model’s ability to approximate human perception of image quality can be achieved.

Our model integrates both traditional and deep learning components. Therefore, in our experimental comparisons, we evaluated against representative methods from both categories—feature engineering and deep learning approaches. Specifically, we compared against PSNR \cite{gonzalez2009digital}, SSIM \cite{wang2004image}, VIF \cite{sheikh2005visual}, DISTS \cite{ding2020image}, and LPIPS \cite{zhang2018unreasonable}.

\subsection{Results evaluation}\label{results_evaluation}
Figure \ref{fig:visual_compare} presents two examples from the test split of the 2AFC subset. As described earlier, each example includes three images: $image0$, $image1$, and a $reference$ image. Each algorithm or model (ours and the baselines) assigns quality scores to $image0$ and $image1$ relative to the reference. These relative scores are then compared to the human judgment (ground truth) to evaluate their correctness.

In the example shown in Figure \ref{fig:visual_compare}(a), the human visual system clearly ranks the images as $image0 > reference > image1$, indicating that $image0$ has higher quality than $image1$. The checkmarks below Figure 1(a) represent the decisions made by each algorithm or model regarding the relative quality of $image0$ and $image1$. All baseline methods incorrectly rank $image1$ as having higher quality than $image0$, contrary to human perception. In contrast, our model aligns with the human visual judgment.

Similarly, in Figure \ref{fig:visual_compare}(b), the human visual system can easily discern that $image1$ has significantly better quality than $image0$, with the correct ranking being $image1 > reference > image0$. However, all baseline algorithms and models incorrectly rank $image0$ as having higher quality than $image1$, contradicting human perception. In contrast, our model produces a judgment aligned with human visual assessment, demonstrating that in this example, SPIPS more closely reflects the human visual system.
\subsection{Ablation studies}

To evaluate the impact of the newly introduced components in our model, we conducted an ablation study. Specifically, we designed two ablation scenarios:

\begin{figure*}[t]
    \centering
    \setlength{\tabcolsep}{2pt}
    \begin{tabular}{ccccccc}
        \begin{overpic}[width=\resLen]{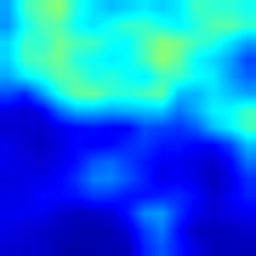}
            \put(18, 3){\scriptsize {\color{white} SPIPS0: $0.236$}}
        \end{overpic} 
        &
        \begin{overpic}[width=\resLen]{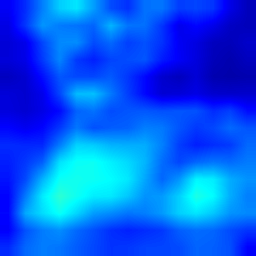}
            \put(18, 3){\scriptsize {\color{white} SPIPS1: $0.202$}}
        \end{overpic} 
        &
        \begin{overpic}[width=\resLen]{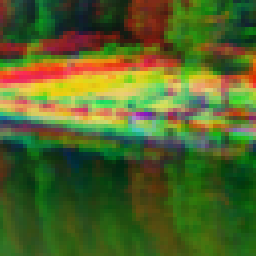}
            \put(35, 3){\scriptsize {\color{white} image0}}
        \end{overpic}
        &
        \begin{overpic}[width=\resLen]{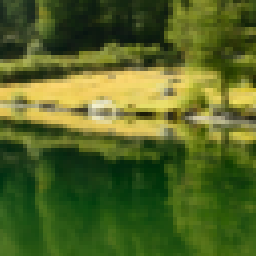}
            \put(28, 3){\scriptsize {\color{white} reference}}
        \end{overpic} 
        &
        \begin{overpic}[width=\resLen]{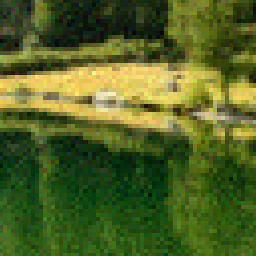}
            \put(35, 3){\scriptsize {\color{white} image1}}
        \end{overpic} 
        &
        \begin{overpic}[width=\resLen]{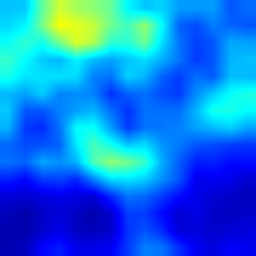}
            \put(18, 3){\scriptsize {\color{white} ablation0: $0.244$}}
        \end{overpic}
        &
        \begin{overpic}[width=\resLen]{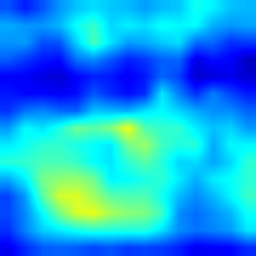}
            \put(18, 3){\scriptsize {\color{white} ablation1: $0.298$}}
        \end{overpic} 
    \end{tabular}
    \caption{Qualitative comparison between the full SPIPS model and its ablated variant without the semantic module. \textit{SPIPS0} and \textit{SPIPS1} denote the evaluation scores assigned by the SPIPS model to $image0$ and $image1$, respectively. The images $image0$ and $image1$ are the two candidates being compared, with $reference$ serving as the ground truth. $ablation0$ and $ablation1$ represent the evaluation scores for $image0$ and $image1$ produced by an ablated version of the SPIPS model, which excludes the semantic module. \textbf{Human preference}: $image0 < image1$}
    \label{fig:ablation1_comparison}
\end{figure*}

\begin{figure*}[t]
    \centering
    \setlength{\tabcolsep}{2pt}
    \begin{tabular}{ccccccc}
        \begin{overpic}[width=\resLen]{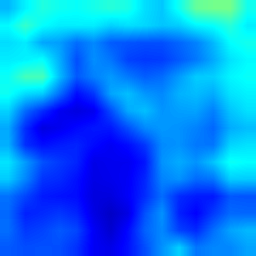}
            \put(18, 3){\scriptsize {\color{white} SPIPS0: $0.246$}}
        \end{overpic} 
        &
        \begin{overpic}[width=\resLen]{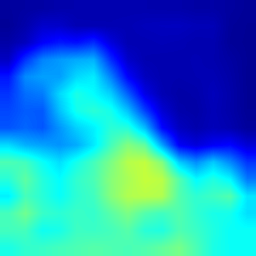}
            \put(18, 3){\scriptsize {\color{gray} SPIPS1: $0.253$}}
        \end{overpic} 
        &
        \begin{overpic}[width=\resLen]{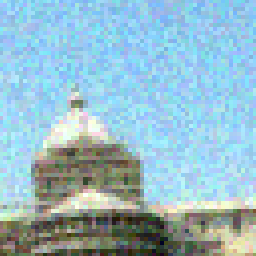}
            \put(35, 3){\scriptsize {\color{white} image0}}
        \end{overpic}
        &
        \begin{overpic}[width=\resLen]{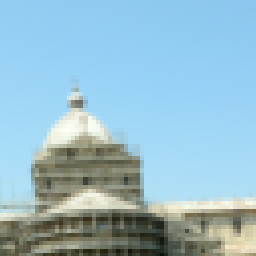}
            \put(28, 3){\scriptsize {\color{white} reference}}
        \end{overpic} 
        &
        \begin{overpic}[width=\resLen]{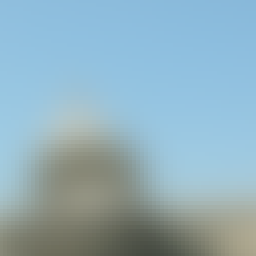}
            \put(35, 3){\scriptsize {\color{white} image1}}
        \end{overpic} 
        &
        \begin{overpic}[width=\resLen]{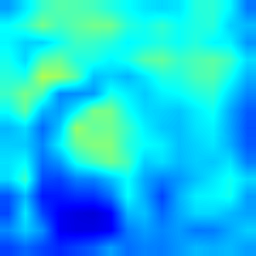}
            \put(18, 3){\scriptsize {\color{white} ablation0: $0.331$}}
        \end{overpic}
        &
        \begin{overpic}[width=\resLen]{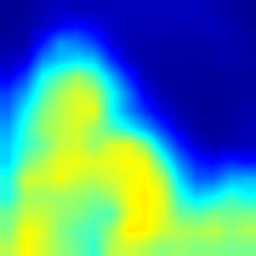}
            \put(18, 3){\scriptsize {\color{gray} ablation1: $0.303$}}
        \end{overpic} 
    \end{tabular}
    \caption{Qualitative comparison of the full SPIPS model and its ablated variant without traditional IQA metrics such as PSNR and SSIM. The roles of \textit{SPIPS0}, \textit{SPIPS1}, $image0$, $reference$, and $image1$ are consistent with those in Figure \ref{fig:ablation1_comparison}. Similarly, $ablation0$ and $ablation1$ represent the evaluation scores of $image0$ and $image1$ generated by the ablated version of the SPIPS model. Unlike Figure \ref{fig:ablation1_comparison}, however, the ablation variant used here excludes traditional IQA metrics. \textbf{Human preference}: $image0 > image1$}
    \label{fig:ablation2_comparison}
\end{figure*}

Ablation 1: Removing the semantic module

Ablation 2: Excluding traditional IQA metrics such as PSNR and SSIM

For each ablation, we trained a corresponding model and evaluated it on the same dataset as the full model. 

Figure \ref{fig:ablation1_comparison} presents a comparison of image quality evaluations produced by the SPIPS model and its ablated variant, \textit{Ablation 1}. Since the \textit{Ablation 1} model excludes the semantic feature module, it lacks the SPIPS model’s ability to assess overall image structure and instead emphasizes local details. In this example, $image0$ appears smoother or lower in resolution, while $image1$ has higher resolution but also more noise. Because \textit{Ablation 1} places greater weight on noise when evaluating image quality, it rates $image1$ as lower in quality compared to $image0$.

The feature maps in Figure \ref{fig:ablation1_comparison} illustrate the differences between the SPIPS and \textit{ablation 1} model in evaluating the quality of $image0$ and $image1$ (with blue indicating low differences and red/yellow indicating high differences). The lower halves of $image0$, $reference$, and $image1$ primarily contain homogeneous content (e.g., a water surface). In the \textit{SPIPS0} feature map, this region shows minimal difference between $image0$ and the $reference$. In the \textit{SPIPS1} map, the difference between $image1$ and the $reference$ in the same region is slightly greater—reflected by a blue-green hue—yet still relatively low. However, the \textit{ablation1} feature map displays green-yellow tones in the lower region, suggesting that the \textit{ablation 1} model perceives a significant difference between $image1$ and the $reference$. This contrasts with human perception, which would consider the lower halves of $image1$ and the $reference$ to be visually similar. The discrepancy arises because $image1$ contains more noise and fine details, which the \textit{ablation 1} model interprets as dissimilarity, whereas human observers tend to overlook such minor variations and view $image1$ and the $reference$ as more alike.

Figure \ref{fig:ablation2_comparison} presents the comparison between the SPIPS model and the \textit{Ablation 2} variant. Unlike the \textit{Ablation 1} model, \textit{Ablation 2} prioritizes semantic and perceptual features while de-emphasizing fine image details. As a result, it rates $image1$ higher in quality than $image0$. This is reflected in its amplification of the structural differences in the tower region of $image1$ and the $reference$, while downplaying the subtle detail differences in more homogeneous areas, such as the sky.

Statistical results are presented in Tables \ref{tab:ablation_plcc}, \ref{tab:ablation_srcc}, and \ref{tab:ablation_krcc}, which compare the performance of the two ablated models with that of the complete model.

As expected, both ablation models performed worse than the full model across all metrics. These results validate the effectiveness of the newly added components and reinforce their contribution to the overall performance of the proposed model.

\begin{table}[H]
    \centering
    \small
    \setlength{\tabcolsep}{1.5pt}
    \begin{tabular}{m{1.8cm}*{6}{m{1.0cm}}}
        \toprule
         \textbf{2AFC} & \multicolumn{6}{c}{PLCC$\uparrow$ (Pearson's Linear Correlation Coefficient)} \\ 
         \cmidrule(lr){2-7}
         & CNN & Color & Deblur & Interp & SR & Trad \\
        \midrule
        SPIPS-abla1 & $0.78$ & $0.45$ & $0.36$ & $0.39$ & $0.52$ & $0.67$ \\
        SPIPS-abla2 & $0.79$ & $0.47$ & $0.39$ & $0.42$ & $0.54$ & $0.69$ \\
        SPIPS & $\textbf{0.81}$ & $\textbf{0.51}$ & $\textbf{0.41}$ & $\textbf{0.45}$ & $\textbf{0.59}$ & $\textbf{0.71}$ \\
        \bottomrule
    \end{tabular}
    \captionsetup{format=plain, labelsep=space}
    \captionsetup{font=footnotesize}
    \caption{PLCC comparison of SPIPS model and ablations (2AFC Dataset).}
    \label{tab:ablation_plcc}
\end{table}

\begin{table}[H]
    \centering
    \small
    \setlength{\tabcolsep}{1.5pt}
    \begin{tabular}{m{1.8cm}*{6}{m{1.0cm}}}
        \toprule
        \textbf{2AFC} & \multicolumn{6}{c}{SRCC$\uparrow$ (Spearman's Rank Correlation Coefficient)} \\
        \cmidrule(lr){2-7}
         & CNN & Color & Deblur & Interp & SR & Trad \\
        \midrule
        SPIPS-abla1 & $0.77$ & $0.45$ & $0.36$ & $0.38$ & $0.52$ & $0.67$ \\
        SPIPS-abla2 & $0.78$ & $0.47$ & $0.39$ & $0.41$ & $0.54$ & $0.69$ \\
        SPIPS & $\textbf{0.80}$ & $\textbf{0.51}$ & $\textbf{0.41}$ & $\textbf{0.45}$ & $\textbf{0.59}$ & $\textbf{0.71}$ \\
        \bottomrule
    \end{tabular}
    \captionsetup{format=plain, labelsep=space}
    \captionsetup{font=footnotesize}
    \caption{SRCC comparison of SPIPS model and ablations (2AFC Dataset).}
    \label{tab:ablation_srcc}
\end{table}

\begin{table}[H]
    \centering
    \small
    \setlength{\tabcolsep}{1.5pt}
    \begin{tabular}{m{1.8cm}*{6}{m{1.0cm}}}
        \toprule
        \textbf{2AFC} & \multicolumn{6}{c}{KRCC$\uparrow$ (Kendall's Rank Correlation Coefficient)} \\
        \cmidrule(lr){2-7}
        & CNN & Color & Deblur & Interp & SR & Trad \\
        \midrule
        SPIPS-abla1 & $0.69$ & $0.40$ & $0.32$ & $0.34$ & $0.46$ & $0.60$ \\
        SPIPS-abla2 & $0.70$ & $0.41$ & $0.35$ & $0.36$ & $0.47$ & $0.61$ \\
        SPIPS       & $\textbf{0.71}$ & $\textbf{0.46}$ & $\textbf{0.36}$ & $\textbf{0.40}$ & $\textbf{0.52}$ & $\textbf{0.63}$ \\
        \bottomrule
    \end{tabular}
    \captionsetup{format=plain, labelsep=space}
    \captionsetup{font=footnotesize}
    \caption{KRCC comparison of SPIPS model and ablations (2AFC Dataset).}
    \label{tab:ablation_krcc}
\end{table}

\subsection{Comparisons with prior works}
Figure 1 presents two examples, each containing three images: $image0$, $image1$, and  $reference$ image. In Section B, we highlight the differences between our model and baseline methods in making visual judgments. Building on this intuitive visual comparison, we provide a detailed quantitative analysis of SPIPS and other models in this section.

As introduced in Section \ref{DAES}, the BAPPS dataset categorizes image quality into six groups: \textit{CNN}, \textit{Color}, \textit{Deblur}, \textit{Interp}, \textit{SR}, and \textit{Trad}. Rather than providing a single overall score, we report performance results for each category individually to offer a more comprehensive evaluation.

Table \ref{tab:compare_plcc} presents the Pearson correlation coefficients between the outputs of six different models/algorithms and human judgments across all six categories. Higher values indicate stronger alignment with human perception. As shown, our model consistently achieves the highest correlation across all categories, demonstrating that SPIPS more closely reflects human visual assessment than competing methods.

Tables \ref{tab:compare_srcc} and \ref{tab:compare_krcc} provide similar quantitative comparisons using different correlation metrics. These results are consistent with those in Table \ref{tab:compare_plcc}, further validating the superior performance of our model.

In addition, Table \ref{tab:compare_jnd} presents results on the JND subset of the BAPPS dataset. Since the JND task focuses on identifying whether a perceptible difference exists between two images—without requiring precise numerical judgments—traditional metrics like PLCC are less effective. Instead, rank-based correlation measures such as SRCC and KRCC are more appropriate. As shown in Table \ref{tab:compare_jnd}, our model achieves superior performance over all baselines in SRCC and KRCC, while exhibiting slightly lower PLCC than LPIPS and DISTS on the CNN sub-dataset, demonstrating its strength in capturing fine-grained perceptual differences.

\begin{table}[ht]
    \centering
    \small
    \setlength{\tabcolsep}{1.5pt}
    \begin{tabular}{m{1.8cm}*{6}{m{1.0cm}}}
        \toprule
         \textbf{2AFC} & \multicolumn{6}{c}{PLCC$\uparrow$ (Pearson's Linear Correlation Coefficient)} \\ 
         \cmidrule(lr){2-7}
         & CNN & Color & Deblur & Interp & SR & Trad \\
        \midrule
        PSNR\cite{gonzalez2009digital} & $0.72$ & $0.40$ & $0.32$ & $0.14$ & $0.41$ & $0.19$ \\
        SSIM\cite{wang2004image} & $0.71$ & $0.38$ & $0.28$ & $0.16$ & $0.35$ & $0.34$ \\
        VIF\cite{sheikh2005visual} & $0.62$ & $0.04$ & $0.32$ & $0.32$ & $0.44$ & $0.16$ \\
        DISTS\cite{ding2020image} & $0.75$ & $0.37$ & $0.35$ & $0.42$ & $0.56$ & $0.63$ \\
        LPIPS\cite{zhang2018unreasonable} & $0.78$ & $0.38$ & $0.35$ & $0.42$ & $0.58$ & $0.61$ \\
        SPIPS (ours)& $\textbf{0.81}$ & $\textbf{0.51}$ & $\textbf{0.41}$ & $\textbf{0.45}$ & $\textbf{0.59}$ & $\textbf{0.71}$ \\
        \bottomrule
    \end{tabular}
    \captionsetup{format=plain, labelsep=space}
    \captionsetup{font=footnotesize}
    \caption{PLCC comparison: SPIPS vs. previous FR-IQA metrics (2AFC dataset)}
    \label{tab:compare_plcc}
\end{table}

\begin{table}[ht]
    \centering
    \small
    \setlength{\tabcolsep}{1.5pt}
    \begin{tabular}{m{1.8cm}*{6}{m{1.0cm}}}
        \toprule
        \textbf{2AFC} & \multicolumn{6}{c}{SRCC$\uparrow$ (Spearman's Rank Correlation Coefficient)} \\
        \cmidrule(lr){2-7}
         & CNN & Color & Deblur & Interp & SR & Trad \\
        \midrule
        PSNR\cite{gonzalez2009digital} & $0.72$ & $0.40$ & $0.31$ & $0.14$ & $0.41$ & $0.19$ \\
        SSIM\cite{wang2004image} & $0.71$ & $0.38$ & $0.27$ & $0.16$ & $0.35$ & $0.34$ \\
        VIF\cite{sheikh2005visual} & $0.61$ & $0.04$ & $0.31$ & $0.32$ & $0.44$ & $0.17$ \\
        DISTS\cite{ding2020image} & $0.75$ & $0.37$ & $0.34$ & $0.41$ & $0.56$ & $0.63$ \\
        LPIPS\cite{zhang2018unreasonable} & $0.77$ & $0.38$ & $0.35$ & $0.41$ & $0.58$ & $0.61$ \\
        SPIPS (ours)& $\textbf{0.80}$ & $\textbf{0.51}$ & $\textbf{0.41}$ & $\textbf{0.45}$ & $\textbf{0.59}$ & $\textbf{0.71}$ \\
        \bottomrule
    \end{tabular}
    \captionsetup{format=plain, labelsep=space}
    \captionsetup{font=footnotesize}
    \caption{SRCC comparison: SPIPS vs. previous FR-IQA metrics (2AFC dataset)}
    \label{tab:compare_srcc}
\end{table}

\begin{table}[ht]
    \centering
    \small
    \setlength{\tabcolsep}{1.5pt}
    \begin{tabular}{m{1.8cm}*{6}{m{1.0cm}}}
        \toprule
         \textbf{2AFC} & \multicolumn{6}{c}{KRCC$\uparrow$ (Kendall's Rank Correlation Coefficient)} \\
         \cmidrule(lr){2-7}
         & CNN & Color & Deblur & Interp & SR & Trad \\
        \midrule
        PSNR\cite{gonzalez2009digital} & $0.65$ & $0.35$ & $0.28$ & $0.12$ & $0.37$ & $0.17$ \\
        SSIM\cite{wang2004image} & $0.64$ & $0.34$ & $0.24$ & $0.14$ & $0.31$ & $0.30$ \\
        VIF\cite{sheikh2005visual} & $0.55$ & $0.03$ & $0.28$ & $0.28$ & $0.39$ & $0.15$ \\
        DISTS\cite{ding2020image} & $0.67$ & $0.33$ & $0.31$ & $0.36$ & $0.50$ & $0.56$ \\
        LPIPS\cite{zhang2018unreasonable} & $0.69$ & $0.34$ & $0.31$ & $0.37$ & $\textbf{0.52}$ & $0.54$ \\
        SPIPS (ours) & $\textbf{0.71}$ & $\textbf{0.46}$ & $\textbf{0.36}$ & $\textbf{0.40}$ & $\textbf{0.52}$ & $\textbf{0.63}$ \\
        \bottomrule
    \end{tabular}
    \captionsetup{format=plain, labelsep=space}
    \captionsetup{font=footnotesize}
    \caption{KRCC comparison: SPIPS vs. previous FR-IQA metrics (2AFC dataset).}
    \label{tab:compare_krcc}
\end{table}

\begin{table}[H]
    \centering
    \small
    \setlength{\tabcolsep}{1.5pt}
    \begin{tabular}{cccc|ccc}
        \toprule
         \textbf{JND} & \multicolumn{3}{c}{CNN} & \multicolumn{3}{c}{Trad}\\[1pt] 
         & PLCC$\uparrow$ & SRCC$\uparrow$ & KRCC$\uparrow$ & PLCC$\uparrow$ & SRCC$\uparrow$ & KRCC$\uparrow$ \\
        \midrule
        PSNR\cite{gonzalez2009digital} & $0.61$ & $0.63$ & $0.50$ & $0.16$ & $0.10$ & $0.07$ \\
        SSIM\cite{wang2004image} & $0.46$ & $0.59$ & $0.46$ & $0.27$ & $0.28$ & $0.22$ \\
        VIF\cite{sheikh2005visual} & $0.53$ & $0.56$ & $0.44$ & $0.13$ & $0.07$ & $0.05$ \\
        DISTS\cite{ding2020image} & $\textbf{0.63}$ & $0.67$ & $0.53$ & $0.01$ & $0.05$ & $0.03$ \\
        LPIPS\cite{zhang2018unreasonable} & $\textbf{0.63}$ & $0.71$ & $0.56$ & $0.55$ & $0.57$ & $0.45$ \\
        SPIPS (ours)& $0.60$ & $\textbf{0.73}$ & $\textbf{0.58}$ & $\textbf{0.55}$ & $\textbf{0.65}$ & $\textbf{0.52}$ \\
        \bottomrule
    \end{tabular}
    \captionsetup{format=plain, labelsep=space}
    \captionsetup{font=footnotesize}
    \caption{PLCC, SRCC and KRCC comparisons: SPIPS vs. previous FR-IQA metrics (JND dataset)}
    \label{tab:compare_jnd}
\end{table}

\section{conclusion}
Motivated by the growing adoption of deep learning-based image quality assessment (IQA) methods in recent years, we have also observed inconsistencies among various IQA metrics and noticeable discrepancies between the assessments made by computer vision algorithms and those made by the human visual system. These observations inspired us to explore a more unified and human-aligned approach to image quality evaluation.

In this work, we propose a unified framework that integrates both traditional and deep learning-based IQA methods. Our approach extracts and processes semantic features separately—capturing high-level image content—before combining them with low-level perceptual features and traditional IQA metrics. This integration aims to offer a more comprehensive and objective evaluation of image quality that aligns more closely with human visual perception.

Initial results are promising. Looking ahead, our future work will focus on two main directions. First, we plan to expand our dataset with more diverse and perceptually meaningful samples to enhance the model’s understanding of image quality. Second, we aim to incorporate vision transformer architectures to further refine our model. By generating tokens from patches across different feature maps and modeling inter-token relationships, we hope to achieve a deeper and more cognitively aligned understanding of image quality, better reflecting human perceptual habits.

We are optimistic about the continued progress of this research and look forward to sharing further advancements in the near future.

\bibliographystyle{IEEEtran}

\vspace{12pt}

\end{document}